\def\year{2020}\relax
\documentclass[letterpaper]{article} 
\usepackage{aaai20}  
\usepackage{times}  
\usepackage{helvet} 
\usepackage{courier}  
\usepackage[hyphens]{url}  
\usepackage{graphicx} 
\usepackage{graphicx}  
\usepackage{amsfonts,amssymb}
\usepackage{color}
\usepackage{bm}
\usepackage{multirow}

\def\ie{\emph{i.e.}}

\title{Symbiotic Attention with Privileged Information for \\
Egocentric Action Recognition }
\author{
\textbf{Xiaohan Wang}\textsuperscript{\rm 1,2},
\textbf{Yu Wu}\textsuperscript{\rm 1,2},
\textbf{Linchao Zhu}\textsuperscript{\rm 1},
\textbf{Yi Yang}\textsuperscript{\rm 1}\thanks{{This work was done when Xiaohan Wang and Yu Wu interned at Baidu Research. Yi Yang is the corresponding author.}}\\ 
\textsuperscript{\rm 1}ReLER, University of Technology Sydney \ \textsuperscript{\rm 2}Baidu Research\\ 
\{xiaohan.wang-3,yu.wu-3\}@student.uts.edu.au\\
\{linchao.zhu,yi.yang\}@uts.edu.au
}
 
\begin{document}

\maketitle

\begin{abstract}
Egocentric video recognition is a natural testbed for diverse interaction reasoning.
Due to the large action vocabulary in egocentric video datasets, recent studies usually utilize a two-branch structure for action recognition, \ie, one branch for verb classification and the other branch for noun classification. 
However, correlation studies between the verb and the noun branches have been largely ignored. Besides, the two branches fail to exploit local features due to the absence of a position-aware attention mechanism.
In this paper, we propose a novel Symbiotic Attention framework leveraging Privileged information (SAP) for egocentric video recognition. 
Finer position-aware object detection features can facilitate the understanding of actor's interaction with the object. We introduce these features in action recognition and regard them as privileged information.
Our framework enables mutual communication among the verb branch, the noun branch, and the privileged information.
This communication process not only injects local details into global features but also exploits implicit guidance about the spatio-temporal position of an on-going action.
We introduce novel symbiotic attention (SA) to enable effective communication.
It first normalizes the detection guided features on \textit{one branch} to underline the action-relevant information from \textit{the other branch}. SA adaptively enhances the interactions among the three sources. To further catalyze this communication, spatial relations are uncovered for the selection of most action-relevant information.
It identifies the most valuable and discriminative feature for classification. 
We validate the effectiveness of our SAP quantitatively and qualitatively. Notably, it achieves the state-of-the-art on two large-scale egocentric video datasets.

\end{abstract}

\section{Introduction}
We have witnessed a significant progress in tackling many computer vision problems, e.g., image classification~\cite{krizhevsky2012imagenet,he2016deep,huang2017densely}, detection~\cite{girshick2014rich,girshick2015fast,Ren2015FasterRT,He2017MaskR}, segmentation~\cite{long2015fully,chen2017deeplab,He2017MaskR}. In video analysis, with the emerging of deep convolutional neural networks and large-scale datasets, the action recognition performance has been prominently boosted~\cite{simonyan2014two,tran2015learning,zhu2018hidden,TSN2016ECCV,xie2018rethinking,wu2019DAM}. In typical action recognition datasets~\cite{Carreira2017QuoVA,goyal2017something}, the video duration is usually less than 10 seconds.
These trimmed videos contain a single action that dominates the whole clip. As the background environment is not distracting, it is often not necessary to identify the interacting object.
However, in many real-world applications, e.g., a robot navigates in the mall, the surroundings are noisy with multiple actors and objects.

\begin{figure}[t]
\center
\includegraphics[width=1.0\linewidth]{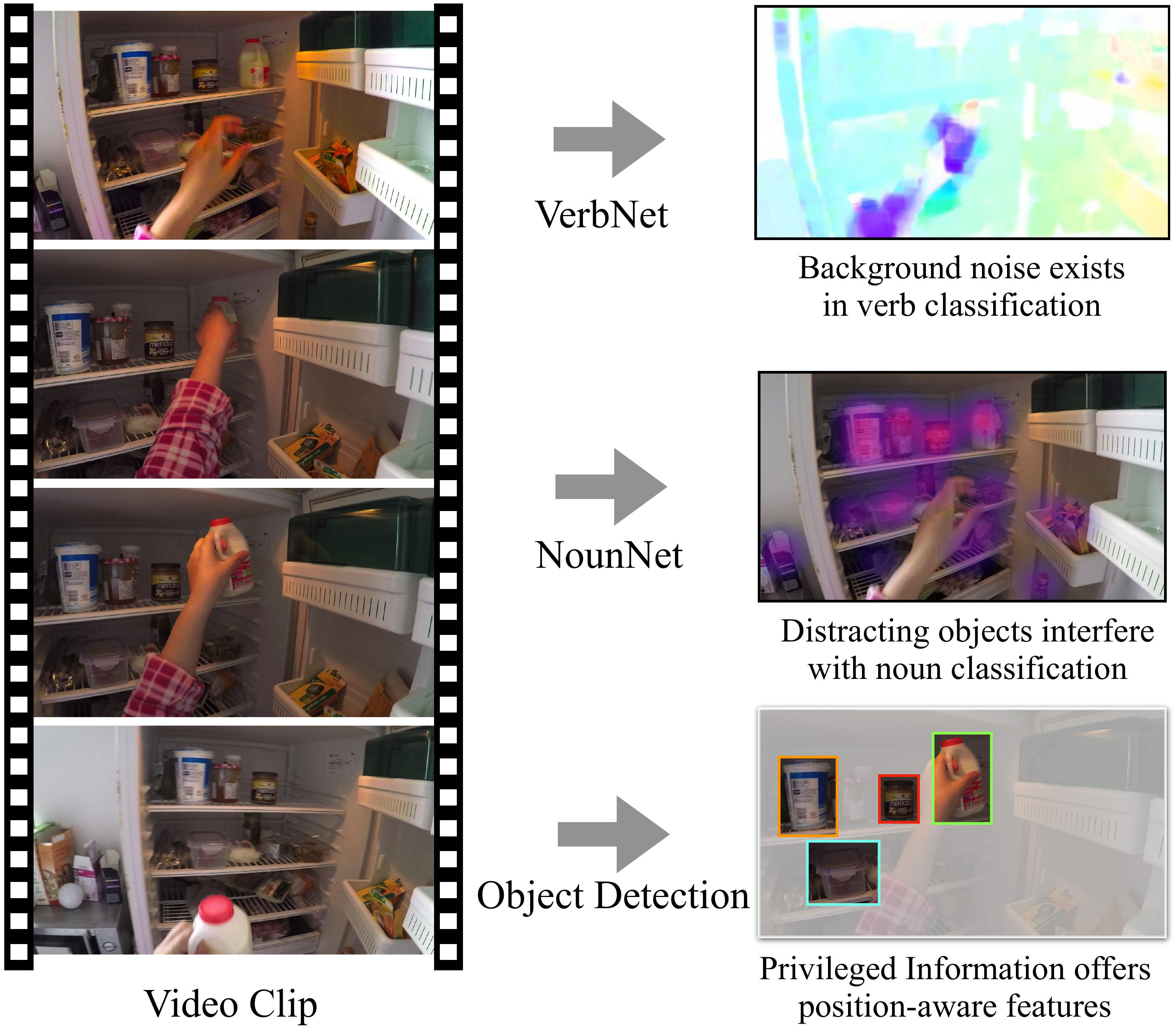}
\caption{
Three sources of information are leveraged. The VerbNet extracts motion information from raw videos, while background noise possibly degrades the recognition of target action.
The NounNet recognizes the object in the scene. However, distracting objects interfere with accurate noun classification.
Local position-aware object detection features serves as privileged information to enhance the communication between two branches. 
}
\label{fig:motivation}  
\end{figure}

Human action recognition in videos has evolved from classifying a single action in a clear background to understanding complex human-object interactions in a highly distracting environment.
To enable the recognition of more complex videos, a challenging large-scale first-person dataset, \textit{i.e.}, EPIC-Kitchens~\cite{Damen2018ScalingEV}, was recently introduced for egocentric daily human activities understanding.
This dataset provides rich interactions, covering adequate objects and natural actions.
Compared to third-person action recognition, it requires to distinguish the object that human is interacting with from
various small distracting objects.
The intense camera motion, occlusion, and first-person viewpoint make it even more challenging to recognize fine actions.

In EPIC-Kitchens, due to the large action vocabulary, 
The verb and the noun classifiers are usually trained separately. The verb branch focuses on classifying verbs, e.g., put, open, that the actor is performing. Large camera motion and subtle occurring action positions are the main obstacles for verb classification. The noun branch is to classify the object the actor is interacting with. As shown in Figure~\ref{fig:motivation}, distracting objects in oblique view decrease the prediction score of the interacting object.
The predictions from the two branches are usually merged without interactions for action classification. 
\cite{Wu2018LongTermFB} utilized 3D convolution neural networks (CNN) for the standalone verb and noun classification.
They leverage object detection features for longer context modeling. However, the long-term feature bank is aggregated via a simple max pooling or average pooling operation, while the more sophisticated non-local operator is found not very effective.
\cite{Baradel2018ObjectLV} introduced object relation network for high-level object reasoning, where the relation modeling facilitates object recognition. However, due to the existence of distracting objects, the learned object relatedness without guided supervision might be not useful to identify the object that human is interacting with.
These works ignore the mutual communication between the standalone branches.
Instead, they only focus on contextual modeling and relation reasoning on a single branch.
Even for a human, it can be difficult to recognize an action by only looking at objects while ignoring the actor's intention, or only understanding motion changes without the awareness of the interacting object. To better exploit the mutual benefits of the interactions among different sources, we make the following contributions.

First, privileged local features are dynamically integrated to encourage the learning of action-relevant representations. 
For verb classification, the negative effect of background noise can be suppressed when target object information is effectively leveraged.
However, noun representation loses finer spatial information. It fails to provide detailed position information to be exploited for the attendance of an on-going action.
Position-aware object detection features offer detailed local understanding of the objects. These features can serve as privileged information to possibly reduce the object-irrelevant motions.
In noun classification, the introduced privileged information enhances the significance of correlated objects. The feature of action-relevant objects will be reinforced, thanks to the finer object presentations from the object detector.

Second, we propose symbiotic attention to enable mutual interactions among the three sources. The privileged information is first merged with the global feature from \textit{one branch}. It is then normalized by a gated channel attention mechanism, which reweights the merged feature. This normalization process underlines the action-relevant information from the feature in \textit{the other branch}. After this, each spatial position has integrated all three sources, \ie, the two branches information, and the privileged information.
We leverage a spatial relation module to further catalyze the communication between the verb and the noun features.
Most action-relevant information is identified to generate a discriminative feature for final classification.
This symbiotic attention mechanism dynamically integrates three sources of information towards better action recognition. 

The effectiveness of our SAP framework is validated quantitatively and qualitatively. We achieve the state-of-the-art performance on two large-scale egocentric video datasets, \ie, EPIC-Kitchens and EGTEA~\cite{li2018eye}. Notably, we outperform the state-of-the-art \cite{Wu2018LongTermFB} by
2.7\% on EPIC-Kitchens test unseen set.

\section{Related Work}
\subsection{Deep Video Recognition}
Deep learning methods have achieved promising performance on the video classification task. \cite{simonyan2014two}  proposed to utilize both RGB frames and optical flow as the 2D CNN input to modeling appearance and motion, respectively. TSN \cite{TSN2016ECCV} extended the two-stream CNN by extracting features from multiple temporal segments.  \cite{tran2015learning} proposed a 3D CNN to learn the spatial-temporal information. I3D  initializes 3D CNN with the inflated weights of 2D CNN.   \cite{Hara2018CanS3} evaluated various 3D CNN architectures on a large-scale video dataset and demonstrated the effectiveness of 3D models. More recently, \cite{xie2018rethinking} and  \cite{tran2018closer}  proposed to decompose the 3D kernel to spatial and temporal convolution. 
Moreover, Recurrent Neural Networks (RNNs) are effective architectures for temporal modeling and have been found useful for video classification in \cite{abu2016youtube,Zhu_2017_CVPR}. 
These deep models are designed for third-person video recognition. They are able to capture motion and scene information but are not sufficient to locate various small objects in egocentric videos accurately.  

\subsection{First Person Action Recognition}
Compared to third-person video recognition, egocentric action recognition is more dependent on the modeling of the hand-object interaction.  \cite{fathi2011understanding} proposed to learn a hierarchical model which exploits the consistent appearance of objects, hands, and actions and refines the object prediction based on action context.  \cite{ma2016going} utilized a hand segmentation net to locate the object of interest. After that, the cropped regions and optical flow images are fed to a two-stream CNN to learn the action and object representation jointly. \cite{Baradel2018ObjectLV} proposed to perform object-level visual reasoning about spatio-temporal interactions in videos through the integration of object detection networks.  More recently, \cite{Wu2018LongTermFB} combined Long-Term Feature Banks that contains object-centric detection features with 3D CNN to improve the accuracy of object recognition. The attention mechanism is efficient to locate the region of interest on the feature map.   \cite{Sudhakaran_2019_CVPR} proposed a Long Short-Term Attention model to focus on features from relevant spatial parts. They extended LSTM with a recurrent attention component and an output pooling component to track the discriminative area smoothly across the video sequence.  \cite{li2018eye} proposed to generate attention map of the hand-object interaction by the guide of the gaze information.  \cite{Kazakos_2019_ICCV} developed a egocentric action recognition model using three modalities.

\subsection{Human-Object Interaction}
Reasoning the interaction between human and objects is relevant to our task. Most methods in this field are based on detection models. For example,  \cite{gkioxari2018detecting} predicted a density map to locate the interacted object and calculated the action score, with a modified Faster RCNN architecture. \cite{qi2018learning} proposed Graph Parsing Neural Networks that incorporates structural knowledge and deep object detection model.  \cite{fang2018pairwise} developed a pairwise body-part attention model which can learn to focus on crucial parts for human-object interaction (HOI) recognition.  
Besides, some works use human-object interactions to help recognize actions.  \cite{wang2018videos} proposed to represent videos as space-time region graphs, which models shape dynamics and relationships between actors and objects.  \cite{sun2018actor} developed an Actor-Centric Relation Network for spatio-temporal action localization.
Most of these HOI techniques rely on the appearance of the actors, which is absent in egocentric videos. Instead of the use of the detection features of humans, we pay attention to the interactions between the motion and the objects.

\section{Proposed Method}

\subsection{Overview}
\begin{figure*}[t]
\center
\includegraphics[width=1.0\linewidth]{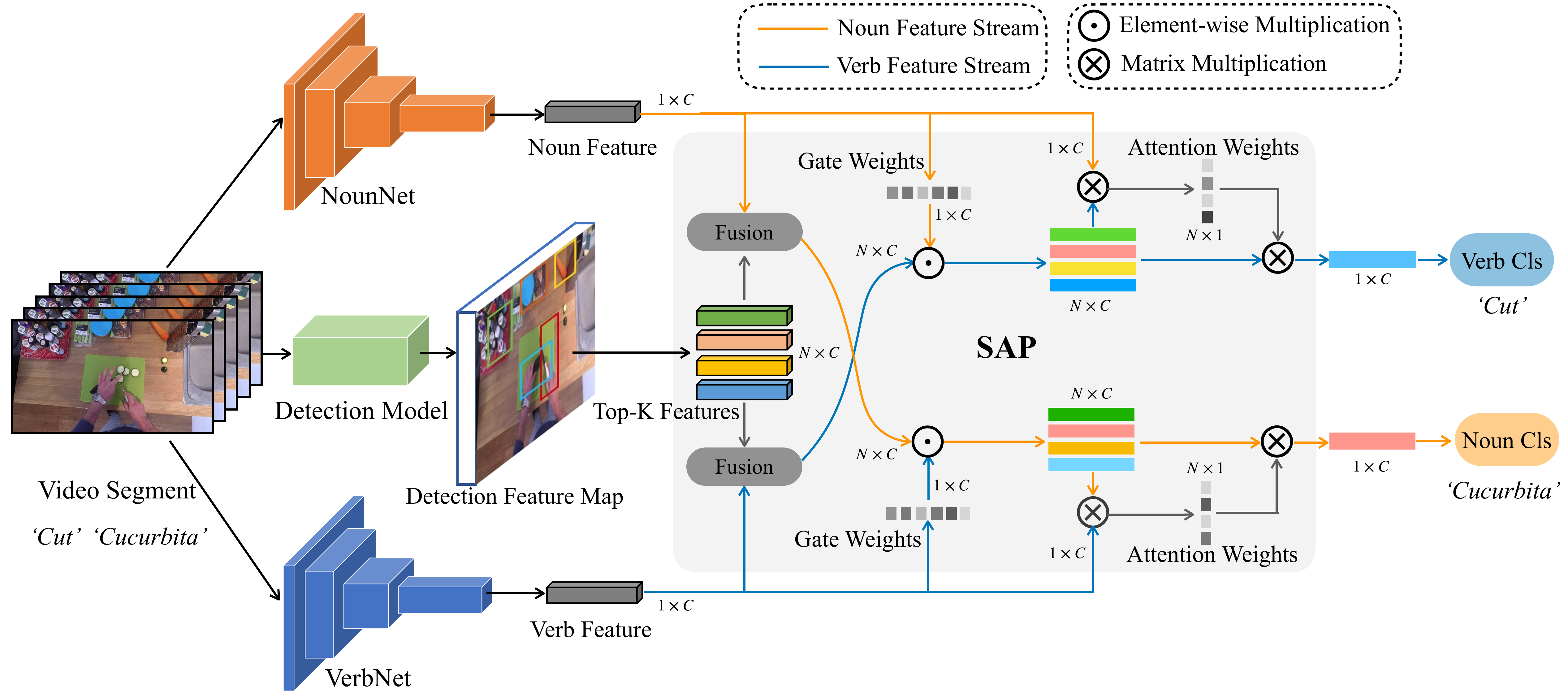}
\caption{
The proposed method. Our framework consists of three feature extractors and one interaction module SAP. 
VerbNet and NounNet produce global features. Detection Model generates a set of local object features as privileged information. 
The privileged information is integrated into the global features to obtain object-centric feature matrices. These feature matrices are normalized by a cross stream gating mechanism. After that, the object-centric matrices are attended by the other branch to select the most action-relevant information. The outputs of SAP are used to classify the verb and noun, respectively.
}
\label{fig:framework}  
\end{figure*}

In this section, we illustrate our network architecture for egocentric video recognition. We develop three base networks to extract features from the input video: 
(1) VerbNet is a 3D CNN and takes a video clip as input. It is designed to capture the motion information. 
(2) NounNet has the same architecture as VerbNet. It is trained to produce a feature representing object appearance. 
(3) Object detection model takes sampled individual frames as input. We use Faster R-CNN as our detector and utilize $RoIAlign$ operation to obtain the local object features as privileged information. The output features of the three base models are fed to the subsequent SAP module. We aim to enable effective communication among VerbNet, NounNet, and Privileged Information. The SAP module generates two feature vectors which can be used to predict verb class and noun class. The overall framework is illustrated in Fig.~\ref{fig:framework}.

\subsection{Preliminaries}
For each input egocentric video ${X} = \{x^1,...,x^t\}$ with $t$ frames,
its verb and noun label is $y^v$ and $y^n$, respectively.
The action $y=(y^v, y^n)$ is a combination of the verb and noun. 
We use two individual 3D CNNs as the backbones in our framework, with one for the verb feature extraction and the other for the noun feature extraction.
The extracted verb feature ${f^v} \in \mathbb{R}^C$ contains the motion information, where $C$ is the dimension of the extracted feature. Differently, the noun feature ${f^n} \in \mathbb{R}^C$ contains the global appearance information.

To enhance the global representation through the communication between two branches, we use a pre-trained detection model to provide detailed information of objects in the video.
Considering the efficiency, for each video, we only use $M$ sampled frames for detection inference.
The output of the $RoIAlign$ layer of the detection model is regarded as the feature for each detected object.
To save memory usage and reduce the noisy information, we only keep top-$K$ object features according to their confidence scores for each sampled frame.
Thus, we have the auxiliary object feature matrix ${f^o}\in \mathbb{R}^{N \times C}$, which contains $N = M\times K$ object features of the video. The verb feature ${f^v}$, noun feature ${f^n}$, and object feature matrix ${f^o}$ are interacted with each other to produce more discriminative features for action recognition with the following SAP module.

\subsection{Symbiotic Attention with Privileged Information}
As illustrated in Fig.~\ref{fig:framework}, SAP includes three stages. First, the privileged information is integrated into the global feature from one branch. Second, the fused object-centric features are recalibrated by the other branch utilizing cross stream gating mechanism. After that, the normalized feature matrix is attended by the other branch to aggregate the most action-relevant information within an action-attended relation module. Considering the symmetry of SAP, we only formulize the noun branch as an example in the following section. 

\paragraph{Privileged Information Integration.}
The separated verb branch and noun branch produce two feature vectors ${f^v}$ and ${f^n}$ by global average pooling. The local information in these features is indistinct. We aim to leverage object feature matrix ${f^o}$ as privileged information to inject the local details into the global features. Moreover, the position-aware object features can also guide the model to attend salient area. Thus, we need to fuse ${f^o}$ with ${f^v}$ and ${f^n}$, respectively.
Besides, it's necessary to avoid jumbling the object feature matrix ${f^o}$. To these ends, we perform a concatenation operation on the object feature matrix and the broadcasting global feature vectors. After that, we use a nonlinear transformation to enhance the fused feature. Formally, this operation can be presented as follow:
\begin{equation}
f^{\hat{n}} = \texttt{ReLU}(W^n_f{f^n} + W^o_f{f^o}+b_f),
\end{equation}
where $W^n_f,W^o_f\in \mathbb{R}^{C\times C},b_f \in  \mathbb{R}^C$ and  ${f^{\hat{n}}} \in \mathbb{R}^{N\times C}$. Each column in ${f^{\hat{n}}}$ represent a object-centric feature, which integrates the global noun appearance with a explicit local object information. 

\paragraph{Cross Stream Gating.}
The fused object-centric feature matrix contains useful local details. However, due to the existence of inaccurate detection regions, there are quite a few disturbing background noises in the features. To address this problem, we propose a gated channel attention to underline the action relevant information. Furthermore, to enhance the interaction between the verb stream and noun stream, we utilize a cross gating mechanism. For an input noun feature matrix 
${f^{\hat{n}}}$, we generate gating weights for it using the verb feature ${f^v}$:
\begin{equation}
{g^n} = \texttt{Sigmoid}(W^n_g{f^v} + b_g),
\end{equation}
where $W^n_g \in \mathbb{R}^{C\times C}, b_g\in \mathbb{R}^{C}$ and ${g^n} \in \mathbb{R}^C$.
The output is produced by rescaling the noun feature matrix with the gating weights:
\begin{equation}
{f^n_g} = {g^n} \odot {f^{\hat{n}}},
\end{equation}
where ${f^n_g} \in \mathbb{R}^{N\times C}$ and $\odot$ denotes the element-wise multiplication.
After re-calibrating the object-centric noun feature by the verb feature, the action-unrelated noise can be suppressed.
Moreover, the cross gating mechanism enables mutual communication between the two branches, which exploits the correlations of verbs and nouns adaptively.
\paragraph{Action-attended Relation Module}
The calibrated object-centric feature matrix contains the action-relevant information and implicit guidance about the spatio-temporal position of an on-going action. To make full use of the information, we consider uncovering the relationships among the features. First, we propose to assess the relevance between the global feature and position-aware object-centric feature. Second, we sum the object-centric features weighted by the relevance coefficients.
Specifically, we perform attention mechanism on the normalized object-centric noun features $f^n_g$ and the original verb feature $f^v$ by
\begin{equation}
{f^n_a} =\texttt{Softmax}({f^n_g}{f^v}){f^n_g},
\end{equation}
where the final noun feature ${f^n_a} \in \mathbb{R}^{C}$.
Through the interaction of global feature and object-centric features, our model selects the most action-relevant feature for classification.

\subsection{Training and Objectives}
We use Faster R-CNN with ResNeXt-101-FPN backbone as our object detector. Following the training procedure in \cite{Wu2018LongTermFB},
we first pre-train the detector on Visual Genome and then finetune it on EPIC-Kitchens object detection set. For VerbNet and NounNet, we adopt 3D Resnet-50 \cite{Hara2018CanS3} as our backbones. The two nets are both initialized with Kinetics pretrained weights. We first individually train the VerbNet and NounNet with the corresponding Cross-Entropy Loss: $\mathcal{L}^{v}$ and $\mathcal{L}^{n}$. After the base training stage, we cascade our SAP module and fine-tune the entire model in an end-to-end manner. The objective for the fine-tuning is the sum of $\mathcal{L}^{v}$ and $\mathcal{L}^{n}$.

\subsection{Action Re-weighting}

The actions are determined by the pairs of verb and noun. 
The basic method of obtaining the action score is to calculate the multiplication of verb probability and noun probability. 
However, there are thousands of combinations and most verb-noun pairs that do not exist in reality, e.g. ``open the knife''. In fact, there are only 149 action classes that have more than 50 samples in EPIC-Kitchens dataset \cite{Damen2018ScalingEV}. 
Following the approach in \cite{Wu2018LongTermFB}, we 
re-weight the final action probability by a prior, \textit{i.e.}
\begin{equation}
  P({action}=y) = \mu(y^v,y^n) P({verb}=y^v)P({noun}=y^n),
\end{equation}
where $\mu$ is the occurrence frequency of action in training set.

\section{Experiments}

\subsection{Datasets}
We evaluate our method on two large-scale egocentric datasets: EPIC-Kitchens~\cite{Damen2018ScalingEV} and EGTEA~\cite{li2018eye}.

\textbf{EPIC-Kitchens} is the largest dataset in first-person vision so far. It consists of 55 hours of recordings capturing all daily activities in the kitchens. The activities performed are non-scripted, which makes the dataset very challenging and close to real-world data. The dataset contains 39,594 action segments which are annotated with 125 verb classes and 321 noun classes. We split the original training set to new training and validation set following \cite{Baradel2018ObjectLV}. We focus on the recognition task on EPIC-Kitchens, which is to predict the verb, noun, and the combination pair in each video segment.

\textbf{EGTEA} is a large-scale egocentric video dataset which consists of 10,321 video clips annotated with 19 verb classes, 51 noun classes, and 106 action classes. There are no bounding boxes annotations in this dataset. 

\begin{table*}[t]
\centering
\begin{tabular}{|c|c|c|c|c|c|c|c|}
\hline
\multirow{2}{*}{Method} & \multirow{2}{*}{Pre-training} & \multicolumn{2}{c|}{Verbs} & \multicolumn{2}{c|}{Nouns} & \multicolumn{2}{c|}{Actions} \\ \cline{3-8} 
 &  & top-1 & top-5 & top-1 & top-5 & top-1 & top-5 \\ \hline
\multicolumn{8}{|c|}{\textbf{{Validation}}} \\ \hline
ORN~\cite{Baradel2018ObjectLV} & ImageNet & 40.9 & - & - & - & - & - \\ \hline
I3D GFA~\cite{wang2019baidu}& Kinetics+ImageNet & - & - & 34.1 & 60.4 & - & - \\ \hline
R(2+1)D~34 \cite{Ghadiyaram_2019_CVPR}& Kinetics & 46.8 & 79.2 & 25.6 & 47.5 & 15.3 & 29.4 \\ \hline
LFB Max \cite{Wu2018LongTermFB} & Kinetics+ImageNet & 52.6 & 81.2 & 31.8 & 56.8 & 22.8 & 41.1 \\ \hline
Ours Baseline & Kinetics & 54.6 & 80.9 & 23.8 & 45.1 & 19.5 & 36.0 \\ \hline
Ours SAP & Kinetics & \textbf{55.9} & \textbf{81.9} & \textbf{35.0} & \textbf{60.4} & \textbf{25.0} & \textbf{44.7} \\ \hline
\multicolumn{8}{|c|}{\textbf{{Test seen (S1)}}} \\ \hline
TSN RGB \cite{price2019evaluation}& ImageNet & 48.0 & 87.0 & 38.9 & 65.5 & 22.40 & 44.8 \\ \hline
TSN Flow \cite{price2019evaluation}& ImageNet & 51.7 & 84.6 & 26.8 & 50.6 & 16.8 & 33.8 \\ \hline
TSN Fusion \cite{price2019evaluation}& ImageNet & 54.7 & 87.2 & 40.1 & 65.8 & 25.4 & 45.7 \\ \hline
R(2+1)D~34 \cite{Ghadiyaram_2019_CVPR}& Kinetics & 59.1 & 87.4 & 38.0 & 62.7 & 26.8 & 46.1 \\ \hline
LSTA \cite{Sudhakaran_2019_CVPR}& ImageNet & - & - & - & - & 30.2 & - \\ \hline
LFB Max \cite{Wu2018LongTermFB}& Kinetics+ImageNet & 60.0 & \textbf{88.4} & 45.0 & \textbf{71.8} & 32.7 & 55.3 \\ \hline
Ours SAP & Kinetics & \textbf{63.2} & 86.1 & \textbf{48.3} & \textbf{71.5} & \textbf{34.8} & \textbf{55.9} \\ \hline
\multicolumn{8}{|c|}{\textbf{Test Unseen (S2)}} \\ \hline
TSN RGB \cite{price2019evaluation}& ImageNet & 36.5 & 74.4 & 22.6 & 46.9 & 11.3 & 26.3 \\ \hline
TSN Flow \cite{price2019evaluation}& ImageNet & 47.4 & 77.0 & 21.2 & 42.5 & 13.5 & 27.5 \\ \hline
TSN Fusion \cite{price2019evaluation} & ImageNet & 46.1 & 76.7 & 24.3 & 49.3 & 14.8 & 29.8 \\ \hline
R(2+1)D~34 \cite{Ghadiyaram_2019_CVPR}& Kinetics & 48.4 & 77.2 & 26.6 & 50.4 & 16.8 & 31.2 \\ \hline
LSTA \cite{Sudhakaran_2019_CVPR}& ImageNet & - & - & - & - & 15.9 & - \\ \hline
LFB Max \cite{Wu2018LongTermFB}& Kinetics+ImageNet & 50.9 & 77.6 & 31.5 & 57.8 & 21.2 & 39.4 \\ \hline
Ours SAP & Kinetics & \textbf{53.2} & \textbf{78.2} & \textbf{33.0} & \textbf{58.0} & \textbf{23.9} & \textbf{40.5} \\ \hline
\end{tabular}
\caption{The comparison with the state-of-the-art methods on the EPIC-Kitchens dataset.}  
\label{tab:epic_sota}
\end{table*}

\subsection{Experiment Settings}
 We implement and test our method using Caffe2, PaddlePaddle and Pytorch. We observe similar performance from these deep learning frameworks. 
Specifically, we pretrain the base models (VerbNet and NounNet) individually and then fine-tune the entire model in an end-to-end manner. We find in our experiments that end-to-end training consistently yields better performance than two-stage training, which only updates the SAP model at the second stage.
Next, we first illustrate the details on how to pre-train the backbones (Backbone details) and how to extract privileged information (Detection details). Finally, we show the details of the end-to-end fine-tuning (SAP details).

\textbf{Backbone details.} We take the Kinetics pre-trained ResNet50-3D model as the initialization of our backbone model.
We then train the backbone models (VerbNet and NounNet) individually on the target dataset.
The input of the two models is the trimmed video clips with 64 frames. The targets for the VerbNet and NounNet are the verb label and noun label, respectively.
Videos are decoded at 60 FPS for the EPIC-Kitchens dataset, and 24 FPS for the EGTEA dataset.
We adopt the stochastic gradient descent (SGD) with momentum 0.9
and weight decay 0.0001 to optimize the parameters for 40
epochs. The overall learning rate is initialized to 0.003 and then changed to 0.0003 in the last 10 epochs. The batch size is 32.
During training, the frame size is $224\times224$ pixels, randomly cropped from a random scaled video whose side is randomly sampled in [224, 288]. We sample 64 frames with stride=2 for both datasets.
During testing, for each input video segment, we use the center clip and scale it to the size $256\time 256$. Then we use the center cropped frames with size $224\time 224$ as the input.

\textbf{Detection details.} Following~\cite{Wu2018LongTermFB}, we use the same Faster R-CNN to detect objects and extract object features. 
The detector is first pre-trained on Visual Genome~\cite{Krishna2016VisualGC} and then fine-tuned on the training split of the EPIC-Kitchens dataset.
For fine-tuning, we use a batch size of 12 and train the model for 15k iterations.
We use an initial learning rate of 0.005, which is decreased by a factor of 10 at iteration 116k and 133k.
Finally, our object features are extracted using RoIAlign from the detector's feature maps.
Considering the efficiency, for a video clip, we extract object features on the center window with a size of 6 seconds.
The sample rate is two frames per second.
For each frame, we keep the top five features according to the confidence scores. 
Therefore, we have 60 detection features for a video clip. For EGTEA, we use the detection model pretrained on EPIC-Kitchens to extract object features.

\textbf{SAP details.}
With the pre-trained backbone models and the detection results, we fine-tune the backbone models with our SAP module in an end-to-end manner.
We use the same SGD to optimize the parameters for 40 epochs with batch-size of 32. The learning rate is initialized to 0.0001 and then reduced to 0.00001 in the last 20 epochs. The rest training details are the same as the backbone details.

\subsection{Comparison with State-of-the-art Results}
We compare our model with the following state-of-the-art methods. \textbf{\textit{TSN}}~\cite{price2019evaluation} is a two-stream model for video recognition. The performance is provided by the dataset authors. \textbf{\textit{ORN}}~\cite{Baradel2018ObjectLV} introduces object relation reasoning upon detection features, while the interactions between verb and noun branches are largely ignored.
\textbf{\textit{LFB}}~\cite{Wu2018LongTermFB} combines Long-Term Feature Banks (detection features) with 3D CNN to improve the accuracy of object recognition.
``LFB Max'' denotes their best operation in EPIC-Kitchens, which leverages max pooling for feature bank aggregation.
\textbf{\textit{LSTA}}~\cite{Sudhakaran_2019_CVPR} is an attention-based method, they only report the top-1 action accuracy on the test set.
Table~\ref{tab:epic_sota} summarizes the top-1 and top-5 accuracy for verb, noun, and action predictions on the {EPIC-Kitchens} dataset. 

Our model outperforms the state-of-the-art methods by a large margin on all three evaluation splits, \ie, the validation set, the test seen (S1) set and the test unseen (S2) set. 
On the validation set, compared to our baseline model (``Ours Baseline''), our SAP on the noun prediction significantly improves the top-1 accuracy from 23.8\% to 35.0\%. 
Compared to ``LFB Max'', which also utilizes the detection features, our method outperforms them by 3.2\% at top-1 accuracy. 
For the verb prediction, our SAP obtains 1.3\% (from 54.6\% to 55.9\%) top-1 accuracy improvement compared to our baseline model. 
For the final action classification, our method achieves 25.0\% top-1 accuracy, which is higher than ``LFB Max'' by 2.2\%. The significant improvement mainly benefits from the interactions between the verb branch, noun branch, and privileged information.
Similar performance improvement is also observed on the test seen (S1) set and the test unseen (S2) set. 
For the final action prediction, Our SAP outperforms the state-of-the-art method ``LFB Max'' by 2.1\% on S1 and 2.7\% on S2.
Althogh\cite{Ghadiyaram_2019_CVPR} use much more videos (65M videos) and extreme deep 3D CNN to train the model, we still outperform their best model on noun prediction and action prediction on S1. 

The results on the EGTEA dataset is shown in Table~\ref{tab:EGTEA}. The results of Two Stream, I3D and TSN, are provided by the dataset developer \cite{li2018eye}. Ego-RNN \cite{sudhakaran2018attention} and LSTA \cite{Sudhakaran_2019_CVPR} utilize RNNs and attention mechanism for egocentric video recognition. Our method achieves higher accuracy on all splits than the state-of-the-art.

\begin{table}[t]
\centering
\begin{tabular}{|ccc|c|c|}
\hline
\multicolumn{3}{|c|}{Inputs} & \multirow{2}{*}{Methods} & \multirow{2}{*}{Top-1}  \\ \cline{1-3}
Noun & Verb & Pril         &  &  \\ \hline
\checkmark & - & -              & CNN Baseline  & 23.8 \\ \hline
\checkmark & \checkmark &   -   &  Noun + Verb  & 24.2   \\ \hline
- & - &  \checkmark      & Avg Pooling & 24.5  \\ \hline
- & - &  \checkmark      & Max Pooling & 25.6   \\ \hline
 - & \checkmark &  \checkmark      & Ours w/o CSG \& ARM & 30.4  \\ \hline
\checkmark & \checkmark &  \checkmark      & Ours w/o Gating & 33.6  \\ \hline
\checkmark & \checkmark &  \checkmark      & Ours w/o Cross Stream & 33.2  \\ \hline
\checkmark & \checkmark &  \checkmark      & Ours w/o CSG & 32.6  \\ \hline
\checkmark & \checkmark &  \checkmark       & Ours w/o ARM & 32.7   \\ \hline
\checkmark & \checkmark &  \checkmark      & Ours & \textbf{35.0}   \\ \hline
\end{tabular}
\caption{Ablation Study based on noun prediction on the EPIC-Kitchens validation set. ``Pril'' denotes the privileged information from finer object detection features, ARM denotes the action-attended relation module, CSG denotes the cross stream gating.}
\label{tab:ablation}
\end{table}

\begin{table}[t]
\centering
\begin{tabular}{|c|c|c|c|c|}
\hline
Methods & Split1 & Split2 & Split3 & Average \\ \hline
Two Stream & 43.8 & 41.5 & 40.3 & 41.8 \\ \hline
I3D & 54.2 & 51.5 & 49.4 & 51.7 \\ \hline
TSN & 58.0 & 55.0 & 54.8 & 55.9 \\ \hline
Ego-RNN  & 62.2 & 61.5 & 58.6 & 60.8 \\ \hline
LSTA  & - & - & - & 61.9 \\ \hline
Ours & \textbf{64.1} & \textbf{62.1} & \textbf{62.0} & \textbf{62.7} \\ \hline
\end{tabular}
\caption{The comparison with the state-of-the-art methods on the EGTEA dataset.}
\label{tab:EGTEA}
\end{table}

\begin{figure*}[t]
\center
\includegraphics[width=1.0\linewidth]{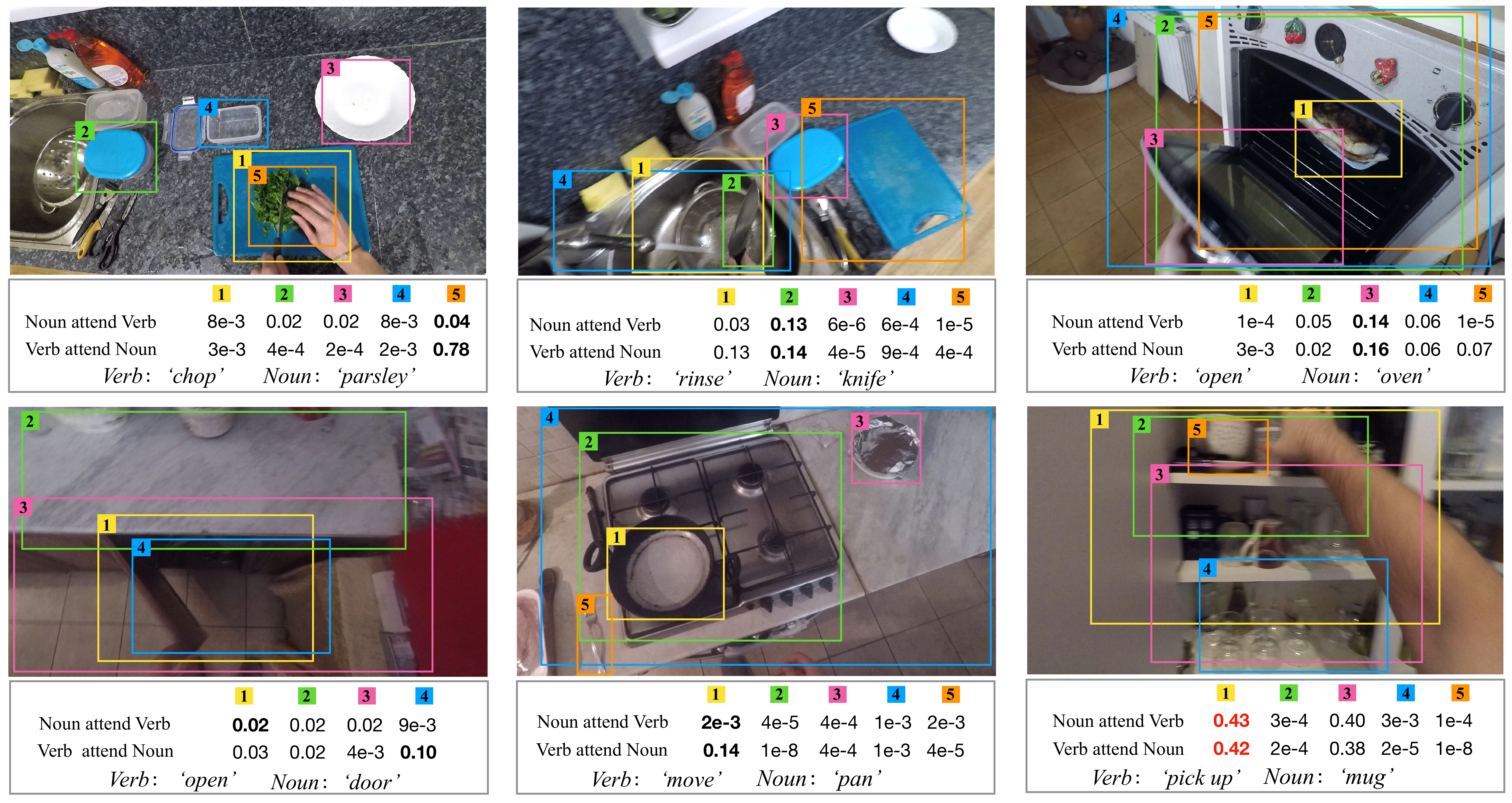}
\caption{
Qualitative results of our SAP model. The colored boxes show the top-5 detected regions and the numbers are the corresponding attention weights generated by our action-attended relation module. Red indicates the failure case.
}
\label{fig:visual}  
\end{figure*}

\subsection{Ablation Studies}
We conduct extensive ablation studies to evaluate the effectiveness of each component in our SAP model. 
Table~\ref{tab:ablation} shows the \textit{noun} prediction accuracy of several variants of our method.
The first row ``CNN Baseline'' indicates we only use the noun branch, without the help of the verb and detection feature.
``Noun + Verb'' is the model that we take the verb feature as the gate to enhance the noun feature. Specifically, the implementation of the gating operation is the same as the single branch in CSG.
``Avg Pooling'' and ``Max Pooling'' take only the privileged information as input. We apply the corresponding pooling operation on the $M*N$ detection features and use a fully connected layer to output the noun prediction. 
The last six rows show the effectiveness of our component CSG and ARM. Specifically, we decompose CSG to two part: Cross Stream and Gating. We also investigate the impact of each part.
In the table, ``Ours w/o Cross Stream'' indicates using the same stream to gate and attend the object-centric matrix.

\textbf{Importance of the privileged information.}
The first two rows in Table~\ref{tab:ablation} shows the importance of the privileged information guidance. 
Without the detection feature, both the noun branch (``CNN Baseline'') and the two branches with communication of noun and verb branches (``Noun + Verb'') fail in the noun prediction.
It is consistent with our motivation that the local information provided by the detection features is a critical clue in the noun prediction.

In addition, the third row (``Avg Pooling'') and the fourth row (``Max Pooling'') also achieve comparable performance compared to the ``CNN Baseline'' model, which indicates the detection features contains rich information for noun prediction. Even without the input of the whole video clip, the local details from the detection model are enough for predicting the interacted object.

\textbf{Importance of Cross Stream Gating.}
The performance comparison between the model ``Ours w/o CSG \& ARM'' and the model ``Ours w/o ARM'' validates the effectiveness of the CSG module.
The CSG module enables mutual communication between the verb branch and the noun branch.
Therefore, the CSG can improve the performance from 30.4\% to 32.7\%. Moreover, the results of ``Ours w/o Cross Stream'' and ``Ours w/o Gating'' demonstrate the two components both benefits the noun prediction.

\textbf{Importance of Action-attended Relation Module.}
The last rows of Table~\ref{tab:ablation} shows the improvement of the proposed ARM. 
ARM can select the most action-relevant information from the object-centric features and explore the relationships in the spatio-temporal context. 
Therefore, the ARM further improves the performance from 32.7\% to 35.0\%.

\subsection{Visualization}
In Fig.~\ref{fig:visual}, we show some qualitative results on the EPIC-Kitchens dataset. 
The colored boxes in the figure indicate the top confident objects found by the pre-trained detection model.
We do not use labels of detected objects since they are not accurate. 
Instead, we use the detection feature to guide the mutual communication of the verb and noun branch. 
The numbers below each image are the value of ARM attention weights for the five object-centric features. 
Taking the first one (the left-top one) as an example, the ground truth of this video clip is ``chop parsley''.
The pre-trained detection model generates five proposals.
Our ARM module correctly finds the interacted area with ``parsley'' and generates a high value (0.78) that describes the contribution of the enhanced feature to the final noun classification.

\section{Conclusion}
In this paper, we propose a novel framework named Symbiotic Attention with Privileged Information for egocentric action recognition. We introduce a new attention mechanism called symbiotic attention that can interactively leverage sources from the verb branch, the noun branch, and the privileged information. 
Our experimental results demonstrate the effectiveness of our framework, and we outperform the state-of-the-art methods on large-scale egocentric video datasets. In the future, we will explore a hierarchical structure that can readily interpret the multi-step attention process. It is also promising to design new models to suppress background distractors directly.

\begin{small}
\bibliographystyle{aaai}
\bibliography{egbib.bib}
\end{small}

\end{document}